# Artificial Empathy Classification: A Survey of Deep Learning Techniques, Datasets, and Evaluation Scales

Sharjeel Tahir
CAIML, School of Science
Edith Cowan University
WA, Australia
s.tahir@ecu.edu.au

Syed Afaq Shah
CAIML, School of Science
Edith Cowan University
WA, Australia

Jumana Abu-Khalaf
CAIML, School of Science
Edith Cowan University
WA, Australia

*Abstract*—From the last decade, researchers in the field of machine learning (ML) and assistive developmental robotics (ADR) have taken an interest in artificial empathy (AE) as a possible future paradigm for human-robot interaction (HRI). Humans learn empathy since birth, therefore, it is challenging to instill this sense in robots and intelligent machines. Nevertheless, by training over a vast amount of data and time, imitating empathy, to a certain extent, can be possible for robots. Training techniques for AE, along with findings from the field of empathetic AI research, are ever-evolving. The standard workflow for artificial empathy consists of three stages: 1) Emotion Recognition (ER) using the retrieved features from video or textual data, 2) analyzing the perceived emotion or degree of empathy to choose the best course of action, and 3) carrying out a response action. Recent studies that show AE being used with virtual agents or robots often include Deep Learning (DL) techniques. For instance, models like VGGFace are used to conduct ER. Semi-supervised models like Autoencoders generate the corresponding emotional states and behavioral responses. However, there has not been any study that presents an independent approach for evaluating AE, or the degree to which a reaction was empathetic. This paper aims to investigate and evaluate existing works for measuring and evaluating empathy, as well as the datasets that have been collected and used so far. Our goal is to highlight and facilitate the use of state-of-the-art methods in the area of AE by comparing their performance. This will aid researchers in the area of AE in selecting their approaches with precision.

## I. INTRODUCTION

The first use of empathy dates back to ancient Greece, when Aristotle was studying how people interact with each other and came up with the idea that we are all connected through a shared consciousness [1]. This theory was later expanded upon by Plato in his book "The Republic", where he described how each person has their own world inside their head, which they can only see through the eyes of others [1]. It can be simply portrayed using the words of the psychologist Theodore Lipps, that states empathy as "feeling one's way into the experience of another" [2].

Identify applicable funding agency here. If none, delete this.

Researchers such as Hoffman [3], Davis [4], and Preston et al. [5] have attempted to unify different perspectives on empathy by adopting a multidimensional approach and by providing comprehensive models of empathy. In particular, Hoffman [3] defines empathy as:

A psychological process that makes a person have "feelings that are more congruent with another's situation than with his own situation."

Empathy is often categorized into three levels: emotional empathy, cognitive empathy and compassionate empathy [6]. Emotional empathy, also known as affective empathy, is the ability to respond to someone's mental state with an appropriate emotion. Cognitive empathy, that overlaps with the concept of "theory of mind", in terms of definition, is the ability to identify and understand others' state of mind. Compassionate empathy is the ability to physically respond to someone's emotional state [7] [8] [9]. When it comes to defining empathy for artificial agents, there are many models of empathy describing it as an innate response, making the im- plementation of empathy on artificial agents complex. For this reason, researchers in HRI, such as [10], [11] have proposed parallel or reactive empathy models. These are synonymous to another aspect of empathy - somatic empathy, which is the ability to spontaneously mimic physical responses, such as facial emotions and physical gestures [12]. Figure 1 shows a schematic description of the different levels of empathy.

It is critical to have an understanding of the ways in which artificial empathy varies from natural or human empathy before moving on to the spectrum of artificial empathy's implementation and its many elements. While in literature AE has been associated with a number of terminologies such as, "empathic computing" [14], "affective computing" [15], and "emotional intelligence" [16], the concept behind all these denotations revolves around the idea of "artificial agents mimicking empathy". In literature, artificial empathy is used more metaphorically, with researchers thinking computers or other artificial agents to have attributes akin to human empathy. This might include descriptions of computers capable of displaying compassion, understanding, or other emotional reactions associated with human empathy. Overall, the use of artificial empathy in literature is often diverse and imaginative, indicating how this subject has caught the imagination of scholars. While these representations are not necessarily scientifically accurate or anchored in real-world technology, they may be a good method to investigate the implications artificial comprehension and reaction to human emotions.

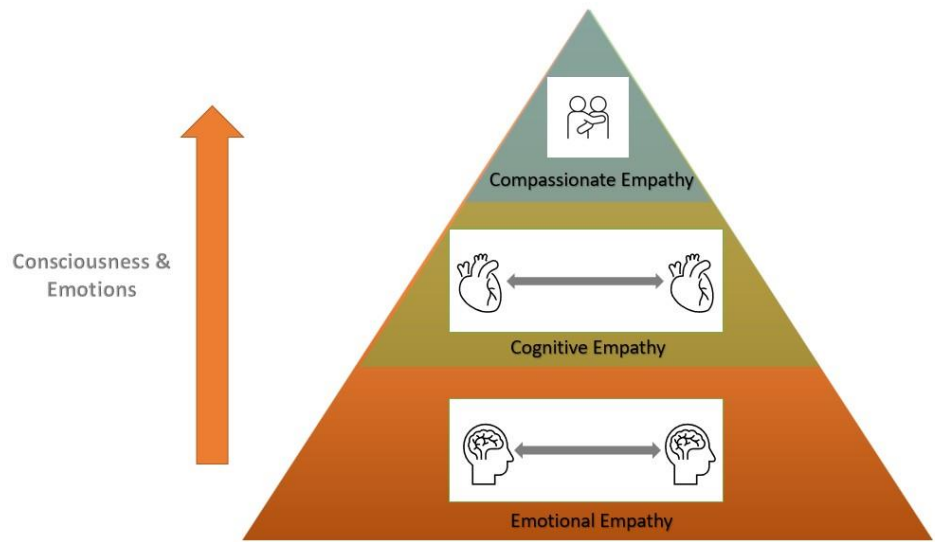

Fig. 1. Levels/Aspects of empathy inspired by Asada et al. [13]

While there are several reasons to endow artificial agents with empathy, it is especially vital because we want our robotic companions to learn what it is like to be human. Building robots with strong empathy skills can benefit the society by allowing machines to comprehend and comfort people in times of distress by picking up on cues about how they are feeling. [17]. This may also assist them in learning how to behave correctly while interacting with people, allowing them to become more effective companions over time [18]. Hence, there has been a lot of interest from the past couple of decades in using technology to make robots more emotionally intelligent — for example, by training them to recognize when someone is upset and then communicate or interact with them in such a way that makes them feel better (or at least less upset, as a human companion would do) [18].

From medical advice tools to companion/social robots, empathy is a crucial aspect of a wide variety of platforms. For instance, people who live in isolation, such as the elderly population or healthcare users who are confined because of infectious diseases such as Covid-19, may benefit from the emotional support that can be provided by online platforms like CRECA (Context Representative Counseling Agent) and emotionally-intelligent companion robots like ARI [19]. CRECA is a classic example of the several online chatbot-based assistants, that have largely been offered as an effective option to aid individuals with mental health difficulties. On the other hand, companion robots offer a wide range of uses, for instance, domestic chores, health monitoring in facilities such as aged care, and providing social assistance to autistic children. While the bulk of these uses require an emotionally competent robot, it is more vital that agents that participate in the delivery of basic medical care, be emotionally capable [19]. Numerous studies have proven how emotional and empathic capabilities of a robot can enhance the experience of human-robot interaction and collaboration [9]. This is mainly because agents with empathy skills are seen as more compassionate and trustworthy than those without, and because they may inspire empathy in their users [20].

The distorted realities of AE:

1) Falsely claiming that current technologies are capable of artificial empathy when they are not.: Existing technologies, such as chatbots or collaborative robots, have been claimed to have "artificial empathy" by some academics, even if they do not entail the understanding or reaction to "real human emotions". For instance, "emotionally intelligent" robots are trained to perform a set number of actions in response to certain human emotions or actions [21]. This may be deceptive since it may lead the user to believe that the technologies are more sophisticated and competent than they really are.

2) Overselling the abilities of artificial agents.: Researchers have asserted that certain cutting-edge technology, such as affective computing systems and companion robots, can read and react to human emotions with remarkable resemblance to the way humans do (anthropomorphism). However, the accuracy and dependability of these systems are generally lacking, and they often rely on a cycle of discourse that may not incorporate the genuine concept of empathy [22].

3) Not thinking about the moral consequences of AE.: The ethical and sociological ramifications of such technologies, i.e., purposely delivering damaging reactions to users of online chatbots for mental health platforms [23], have not been given sufficient attention in previous research focusing on the technological elements of constructing artificial empathy systems. Artificial empathy has the potential to improve people's lives, but it's important to look at the issue from all angles to ensure we do not overlook crucial aspects.

TABLE I
TAXONOMY OF ARTIFICIAL EMPATHY

| Modality | Textual | | | Visual/Multi-modal | | |
|---|---|---|---|---|---|---|
| Emotion/Intent classification | Transformers e.g., BERT | DNN's e.g., Seq2seq | Other AI Algorithms | Transformers e.g., GPT | DNN's, e.g., LSTMs | Other AI Algorithms |
| Dataset | Textual e.g., Empathetic Dialogues | | | Video e.g., OMG-Empathy | | |
| Evaluaition metrics | PPL, EMO ACC, AUC, Distinct-n | | | CCC, Human evaluation surveys, e.g., BLRI, EQ, RoPE | | |

The surveys that have been previously carried out in the field of AE concentrate their attention, for the most part, on the description of empathy and experimental settings of the works that study implementation and evaluation of AE. For example, they explain the particular aspects, such as the cohort of engaged participants and how their responses were compiled and used [6]. The goal of this paper is to review recent trends in the field of AE, especially the works that employ deep learning (DL) classification methods. Therefore, the studies considered in this survey are predominantly from after the advent of DL. Moreover, we analyse the existing metrics for evaluating empathy and highlight the desiderata for future AE evaluation metrics and benchmarks.

To the best of our knowledge, this is the first work to:
- Review existing DL methods used to implement (detect and elicit) AE.
- Analyse the existing scales of empathy and compare their performance with respect to Artificial Empathy.
- Examine the extent to which the existing datasets can be used to train and evaluate AE models.

We have grouped the existing works into sections as per the processes involved in the implementation of AE. Section II provides a review of the state-of-the-art techniques used to model empathy and how each of those studies evaluated their performance. A detailed analysis and overview of the existing scales used for human-based evaluation of AE is presented in Section III. Datasets that have been used to train AE models are analyzed in Section 4. Section V concludes the paper and discusses the possible future research directions.

## II. EXISTING CLASSIFICATION TECHNIQUES OF ARTIFICIAL EMPATHY

### A. Observations

From what we can gather in the literature, most of the state-of-the-art DL-based techniques have dealt with textual data. It's also the case that only the textual works include the use of more recent models like transformers. When compared to other DL approaches like autoencoders and deep CNNs, transformer models such as GPT, GPT-2, BERT, and its variants have shown encouraging results. Therefore, it is reasonable to assume that vision transformers can boost efficiency while dealing with visual data [24]. Furthermore, when using multi-modal features for predicting user empathic reactions, works like [25] and [26] have demonstrated good results, supporting the fusion of different modalities for this task.

In addition, Reinforcement learning (RL), a common learning approach for artificial agents, has been used in various works [27] [28], leading to more realistic performance from the agents [29]. Deep reinforcement learning can improve this, particularly when intrinsic motivation (as opposed to the commonly used extrinsic motivation) is included as a learning objective [30]. Intrinsic motivation could be characterized, in the context of AE agents, as the sense of programming that prioritizes establishing artificial empathy as the agent's primary objective. Extrinsic motivation, on the other hand, involves concentrating on external outcomes, such as evaluation metrics and user experience. By contrasting intrinsic motivation and extrinsic motivation, as tailored by Bagheri et al. [28], the Figure 2 is a depiction of how the function of intrinsic motivation in the development of artificial empathic agents differs from extrinsic motivation. Developers may be better able to construct agents capable of true emotional connection and great user experiences if they prioritise intrinsic motivation and value empathy as a primary function of the system. In order to demonstrate inclusion of intrinsic goals on a state-of- the-art deep reinforcement learning model (Q-learning model), we modify the architecture presented by Sorrentino et al. [27] in Figure 3. The rationale for abstaining from employing solely emotional cues of the user as a means of providing rewards is multifaceted, with the primary objective of enhancing the quality of the human-robot relationship.

Finally, assessment metrics are crucial, especially in the creation of the event response. Many works have employed grammar and language structure-specific metrics including BLEU, BERT, and PPL [31]. Obviously, such approaches fail when evaluating empathy, since the intensity of feelings is more important than the grammar used to express them. As a result, there is a need for autonomous measures that can give more precise task-specific assessment for AE. A good example is the empathy-based autonomous measure proposed by Lee et al. [32] inspired by the EPITOME framework [33]. It is also apparent that approaches using both types of evaluation, i.e., human and automated, have a more robust and comprehensive examination of their outcomes, highlighting the significance of employing both types of evaluation measures. Furthermore, the comparison between human and automated evaluation methods, interestingly, shows variation in results for many works, raising concern over the reliability of automatic metrics. To exemplify this, as depicted in Figure 4, the models presented by [34] report contradicting scores of relevance and fluency, as compared to empathy.

The methodologies which are currently being used for the classification and modelling of AE are analysed in this section. Based on the type of data used, we classify the research into two major groups i.e., textual and visual/multi-modal. The papers are further classified based on the technique they have implemented, i.e., transformer networks, DNNs, and other algorithms. A summary of the AE detection and elicitation techniques is given in Table 2.

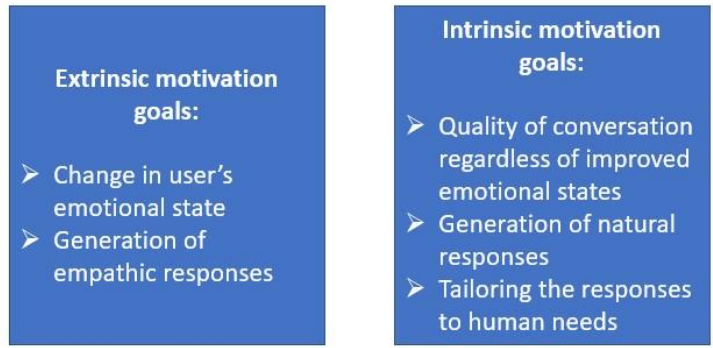

Fig. 2. Examples of intrinsic goals vs extrinsic goals for an artificial empathic agent.

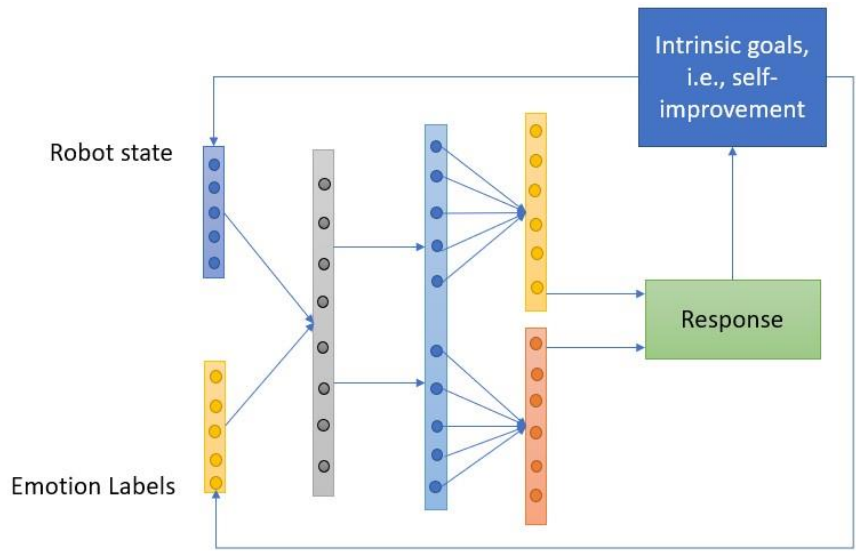

Fig. 3. Modifying the architecture of a Q-learning based RL model originally given by Sorrentino et al. [27] to include intrinsic goals, eliminating human feedback.

|  |  | Empathy | Relevance | Fluency |
|---|---|---|---|---|
| Fine-Tuned | – | $3.25 \pm 0.12$ | $3.33 \pm 0.12$ | $4.30 \pm 0.09$ |
| EmoPrepend-1 | – | $3.16 \pm 0.12$ | $3.19 \pm 0.13$ | $4.36 \pm 0.09$ |
| TopicPrepend-1 | – | $3.09 \pm 0.13$ | $3.12 \pm 0.13$ | $4.41 \pm 0.08$ |
| Fine-tuned | ED | $3.76 \pm 0.11$ | $3.76 \pm 0.12$ | $4.37 \pm 0.09$ |
| EmoPrepend-1 | ED | $3.44 \pm 0.11$ | $3.70 \pm 0.11$ | $4.40 \pm 0.08$ |
| TopicPrepend-1 | ED | $3.72 \pm 0.12$ | $3.91 \pm 0.11$ | $4.57 \pm 0.07$ |

Fig. 4. Conflicting scores of human vs autonomous AE metrics as reported by [34].

## B. State-of-the-art Research in Textual Data

### 1) Transformer based Techniques:

Rashkin et al. [34] proposed a benchmark to evaluate empathetic dialogues/conversation. They presented a dataset that can improve the existing empathetic dialogue generation systems, consisting of 24,850 conversations. The dialogue generation system consists of two modules; Retrieval and Generative. The retrieval-based design is made up of two different transformer encoders; one for the context, and the other for the candidate. BERT [35] serves as the baseline architecture for the encoding of both the context (setting of the dialogue) and the candidates (suitable responses). The whole Transformer architecture [36] is made use of in the generative set-up, where an encoder and a decoder are used. The output of the encoder is used by the Transformer decoder in order to make a prediction about a sequence of words. The Transformer networks employed in their experiments all have the same fundamental architectural makeup (four layers and six attention heads). Performance of the model is evaluated against other techniques using BLEU (Bilingual Evaluation Understudy) and PPL (perplexity) scores. Scores show an improvement when candidate responses are selected from the ED dataset compared to others.

Hosseini et al. [37] took the process of identifying empathy in online chat platforms one step further by introducing a direction of empathy feature - seeking empathy or providing emapathy. They also created a dataset i.e., IEMPATHIZE that consists of 5000 sentences from a cancer platform. The Lexicon-based model consists of emotion lexicon and subjectivity lexicon to establish the baseline's feature set. Strong and weak subjective words were extracted from the conversations and used as features to train a logistic regression model. Extracted TF-IDF feature vectors at the word level are used to train Naive Bayes, Support Vector Machine, and Random Forest, three popular machine learning techniques. Next, they employed a concatenation of CNN, Conv-LSTM, LSTM and Bi-LSTM networks, and fed the outputs of these networks into a fully connected layer for predictions. BERT is also fine-tuned and used as a pre-trained language model. In contrast to the majority of prior text-based empathy methods, they have employed F1 scores to assess the empathy prediction of various models, which is more of an intrinsic measure given that it is not related to language rules, unlike metrics such as BLEU. Results indicate that BERT outperforms other models both with and without pre-training on domain-specific data (more than 5 percent improvement of F1 scores).

In a work by Lee et al. [38], to improve the empathic response generation of ECAs (Embodied Conversational Agents), the Uncertainty-Aware Conditional Variational Auto-Encoder (UACVAE) framework was introduced. A metric called Utterance Entailment (UE) was also included to measure how well an agent's responses fit its circumstances. In their proposed CVAE-based dialogue agent, an approximation of the aleatoric uncertainty of the generated dialogue response is derived from variance of the latent Gaussian distribution. A GPT-2 pretrained model was used to get the sequence embeddings of the dialogue context, the external information, and the generated response. The final loss function integrates the KL divergence between the Gaussian distributions produced by the prior and recognition network. Two variants of the framework have been implemented i.e., UA-CVAE(M) that has several linear layers based combination network and UA-CVAE(C) that has a single layer based combination network. From the combination network, the outputs are fed to the response decoder, which is based on a GPT-2 pretrained language model. In order to train the UA-CVAE, the stochastic gradient variational bayes (SGVB) algorithm is used. The model is implemented on the Empathetic Dialogues dataset [34]. Both automatic and human evaluation were performed, using automatic metrics including PPL, ROGUE [39], METEOR [40], intra-response Distinct-n [41], and their proposed UE score. Results show a strong correlation between the proposed UE score and participant responses. This suggests that UE scores can be used for assessing AE in future studies, however, further experimentation with different datasets and participants would be useful for a complete evaluation of its performance.

Using the newly created psychotherapeutic intervention known as self-attachment technique (SAT), Alazraki et al. [42] offered a computational architecture that augments a rule-based agent for delivery. They compiled 1,181 crowd-sourced emotional utterances and 2,143 sympathetic rewrites of neutral phrases to form a new dataset called Empatheticpersonas. They've implemented a tree-like dialogue flowchart, and they generate innovative yet secure utterances at each node in the chart, with the goal of reducing ambiguity as much as possible. They achieve this by parsing the rewritten utterances in their proposed dataset and removing fragments at important punctuation points to create concise sentences. For the empathy score, they employ a T5 model [43] that has been trained on a labelled subset of the proposed dataset and for the fluency score, they also deduct a penalty for each repeated word inside an utterance from the inverse of its perplexity given by a GPT-2 language model [44]. In addition, they adopt a RoBERTa model [45] for the task of emotion recognition that is trained on an existing affective empathy dataset [46] and further fine-tuned on the expressions of emotion on the Empatheticpersonas dataset. To evaluate the application, human trials with 16 subjects from a non-clinical population, and two medical professionals specialised in mental health, was performed, where subjects were asked to fill out a questionnaire based on questions regarding their interaction with the chatbot. It is worth noting that out of the

TABLE II
SUMMARY OF AE CLASSIFICATION/GENERATION TECHNIQUES

| Study | Technique(s) | Modality | Evaluation metric | Year |
|---|---|---|---|---|
| Rashkin et al. | BERT | Textual | BLEU | 2018 |
| Hosseini et al. | CNN, LSTM, BERT | Textual | F1 scores | 2021 |
| Lee et al. | CVAE, GPT-2 | Textual | PPL, ROGUE, METEOR, Ditinct-n, UE | 2022 |
| Alazraki et al. | RoBERTa, GPT-2 | Textual | Human evaluation | 2021 |
| Sharma et al. | RoBERTa, Bi-encoder | Textual | BLEU, BERT | 2020 |
| Harilal et al. | Seq2seq, LSTM | Textual | BLEU, BERT | 2020 |
| Li et al. | Adversarial NN, CNNs | Textual | PPL, Distinct1 and Distinct2 | 2020 |
| Montiel-Vasquez et al. | PBC4cip, RFM | Textual | CEM, AUC | 2022 |
| Ayshabi et al. | Transformer-Encoder, Decoder | Textual | BLEU, Human evaluation | 2021 |
| Kurashige et al. | RNN | Textual | Human evaluation | 2018 |
| Rasool et al. | CLM, LeaderP clustering, TGART | Textual | Self evaluation | 2015 |
| Lee et al. | GPT-3, RoBERTa | Textual | PPL, Distinct, EPITOME, Human evaluation | 2022 |
| Xie et al. | GPT, SVR | Textual, Speech | PPL, EMO ACC, A.MSE, V.MSE | 2021 |
| Tan et al. | CNN, LSTM | Audio, Textual and Video | Human evaluation, CCC | 2019 |
| Fung et al. | CNN, LSTM | Speech, Audio and Video | Cross-evaluation on datasets | 2018 |
| Mathur et al. | 8 ML models, LSTM, TCN | Video | ACC, AUC, Precision, Recall | 2021 |
| Carolis et al. | STASM, kNN, DBN | Speech and Video | Human evaluation | 2017 |
| Bagheri et al. | SAE | Image | UTAUT, Friendship questionnaire, Engagement parameter | 2020 |
| Filho et al. | RegressionWiSARD | Audio and Video | CCC | 2020 |
| Leite et al. | OPPR, RL | Image | Human evaluation | 2013 |
| Sorrentino et al. | Reinforcement Learning | Video and Audio | UTAUT | 2022 |
| Bagheri et al. | Reinforcement Learning | Video | Godspeed Questionnaire | 2022 |
| Daher et al. | NLTK (Py package) | Textual | RoPe scale | 2022 |
| Roller et al. | GPT3 | Textual | ACUTE-Eval | 2020 |

TABLE III
COMPARISON OF THE MOST COMMONLY USED AUTONOMOUS METRICS FOR EVALUATION OF AE BY STATE-OF-THE-ART METHODS

| Study | Dataset | PPL | Distinct |
|---|---|---|---|
| Rashkin et al. | Empathetic Dialogues | 16.89 | 2.5 |
| Lee et al. | Empathetic Dialogues | 17.16 | 3.1 |
| Sharma et al. | Empathetic Dialogues | 18.32 | - |
| Li et al. | Empathetic Dialogues | 19.09 | 3.0 |
| Harilal et al. | Empathetic Dialogues | 23.60 | 3.3 |
| Ayshabi et al. | Empathetic Dialogues | 23.60 | 3.3 |
| Lee et al. | Empathetic Dialogues | 23.60 | 3.3 |
| Xie et al. | Empathetic Dialogues | 23.60 | 3.3 |

23 participants originally selected, only 16 returned completed evaluation surveys, showing the unreliability of human-based evaluation.

Sharma et al. [33] introduced a conceptual framework (EPITOME) for modelling empathy from online discussions and chats on mental health platforms while introducing a new dataset as well. The framework consisted of three communication mechanisms of empathy: Emotional Reactions (ER), Interpretations (IR), and Explorations (EX). For this purpose, they suggested a RoBERTa-based bi-encoder model to detect empathy-inducing language patterns in conversations. They propose a model based on two independently pre-trained transformer encoders from RoBERTaBASE (S-Encoder & R-Encoder) to encode seeker post and response post, respectively. S-Encoder is responsible for encoding context from the seeker post (the patient or the person who has the enquiry on the web platform), while R-Encoder is in charge of comprehending empathy from the response post. For domain adaptive pre-training of the encoders, they use the Talklife platform [47]. At the last stage of the framework, an empathy identifier module is implemented. It uses the final representation of the seeker's token and compares it with the response's token by passing it through a linear layer to get the predictions. Performance of the model's response is benchmarked using BLEU and BERT scores. In our view, the distinctiveness of their work is the unique approach in which they have labelled the dialogues, i.e., ER, IR, and EX, allowing for improved classification techniques.

In a recent work, Ayshabi et al. [31] proposed a multi-resolution system based on disparate decoders for recognition and processing of emotions and to accumulate the feedback to generate an empathetic response. They used an Emotion-aware Transformer Encoder unit to extract semantic and emotional context from the dialogue. This is then processed and dispatched to the Emotion Expressive Transformer Decoder unit. This results in the identification of the emotion expressed in the dialogue and the generation of a congruent empathetic response. Next, the reaction emotion is learned with the use of separately parameterized decoders, and an empathic response is then generated by a meta decoder that aggregates the weights taken from the decoders. Similar to [38], results are tested on the Empathetic dialogues dataset [48]. Their proposed model was tested against different baseline networks i.e., EmoPrepend-1, MoEL and EmpDG [31] [34], [49], [50], and it proved to perform the best. In addition, to evaluate the model's performance, they used two approaches, i.e., BLEU scores and human evaluation. For human evaluation, participants were asked to evaluate 100 random dialogues against three aspects: empathy, relevancy and fluency.

Lee et al. [32] conducted an empirical study that demonstrated the promising capabilities of GPT-3 in generating empathic responses through in-context learning. Specifically, they trained the model on the Empathetic Dialogues dataset, leveraging emotion information to select in-context examples for training and testing. Their proposed model was benchmarked against two state-of-the-art models, Blender 90M and Emp-GPT3, and was evaluated using a novel autonomous evaluation metric that directly measures empathy. This metric, which is based on the EPITOME framework previously proposed by Sharma et al. [33], measures the difference between empathy scores generated by the model and "human golden responses". In addition, the study used other automatic evaluation metrics, such as PPL and Distinct, to assess the fluency and diversity of the generated responses. Overall, this work provides evidence that GPT-3 can effectively generate empathic responses, and demonstrates the usefulness of an automatic evaluation metric that focuses on the key aspect of empathy, unlike majority of the evaluation works in the area of AE.

Roller et al. [51] used poly-encoder transformer architecture exploiting GPT-3 as the base (using the publicly available ParlAI framework). They highlight the importance of domain-specific training to generate empathic responses. Their proposed model has two disparate parts: retriever and generator, where the retriever selects next dialogues based on candidate scores. The generator, on the other hand, is responsible for generating rather than selecting response from a limited set. The proposed model is then evaluated on the Empathetic Dialogues dataset. Two novel autonomous evaluation metrics for AE are proposed namely TF-IDF and ACUTE-Eval. However, implementation is limited to TF-IDF, as the latter is too costly to implement.

2) Other DNN-based Techniques:

Harilal et al. [52] presented CARO, an empathetic online chatbot which provides support for people suffering of mental health problems. Their proposed technique is an ensemble of two models: one that produces medical advice, and another that generates natural, empathetic dialogue. They introduce an intent feature that decides whether a user should be directed through medical advice model or the empathetic conversation generator. They use separate baseline models i.e., Seq2seq network and an LSTM for empathetic dialog generation and medical advice generation, respectively. Facebook AI Empathetic Dialogue dataset and Medical Q/A dataset are used to train the baseline models. To elicit an empathic reaction, the emotions retrieved by the emotion classifier were attached to the beginning of the context phrase. The model is composed of Encoder-Decoder architecture, where each Encoder and Decoder block is composed of one or more LSTM/GRU units.

Both intent and emotion classification tasks were performed by using an LSTM unit that generates a decoded sequence to be passed on to a dense layer, where Softmax activation is applied for prediction. Performance of the model's response is evaluated using BLEU and BERT scores.

Li et al. [50] proposed a multi-resolution adversarial neural network based model called EmpDG, whose major characteristics are its empathetic generator and its interactive discriminators. The empathic generator uses Transformer-based encoder-decoder design. Semantic context and multi-resolution emotional context are encoded in the encoder; and the decoder combines these to create responses. Two CNN-based discriminators were designed to increase the generator's empathy (i.e., the semantic discriminator and the emotional discriminator). In order to encode the multi-resolution emotional context, they make use of a separate transformer encoder that has a unique set of parameters. As for the discriminators module, the semantic discriminator calculates the semantic distance between the produced and gold responses, the emotional discriminator; determines if the produced reactions are sufficiently empathetic. Both of the discriminators are based on a CNN classifier. Empathetic dialogues dataset is used for experimentation of the model. In contrast to many other studies, they avoid using the BLEU scoring metric for its unreliability as explained by Liu et al. [53]. They use three different automatic evaluation metrics i.e., Perplexity [54], Distinct1 and Distinct2 [41].

Montiel-Vasquez et al. [55] claimed to present the best technique for detecting empathic text, using an explanatory AI model. Their proposed pattern-based classification algorithm [56] - PBC4cip, which is a contrast pattern-based algorithm that is highly capable of performing better in the face of class imbalance problems. The classification algorithms rely on a set of features that serve as representations of the textual data. Textual features (sentiment [57], emotions [58], taxonomy [59], and intent [60]) are extracted using the proposed Paralleldots text mining API for data mining. In addition, relevant features from the Empathetic Conversations are also exploited. Furthermore, pattern extraction from the dataset, which could be used later for explainability purposes, was carried out using the pattern mining algorithm - Random Forrest Miner (RFM) [61]. To further enhance the explicitness of the model, various methods, such as the classifier attribute evaluator [62], and Waikato Environment for Knowledge Analysis (WEKA) [63] are used. The corpus used for training and testing the classification model was the Empathic Dialogues dataset [34]. Five different algorithms—kNN, a Random Forest Classifier (RFC), a Multi-Layer Perceptron (MLP), a Gaussian Naive Bayes (GNB), and a Decision Tree Classifier—were used to evaluate their proposed network's performance. The Closeness Assessment Measure (CEM) and the Area Under the ROC Curve (AUC) [64] were proposed as evaluation measures. Although AUC and CEM are not often used to evaluate empathy, they were helpful here because of the linear trend shown in the scores of empathy.

*3) Other Techniques:*

Kurashige et al. [65] proposed a virtual counselling agent "CRECA" (also implemented on an actual robot) that could help deal with psychological problems of clients through empathic interactions. The architecture of their proposed system consists of two phases namely, problem-discovery and problem-solving. During the phase of problem discovery, the goals are to better understand the nature of the client's challenges and to earn the client's confidence. The client's problem is grouped into 5 or 6 categories after the problem-discovery phase. Thus, problem-solving context is established. During the problem-solving phase, as the client's dialog on specific issues progress, similar keywords are matched single or several times, and replies to deepen the client's contemplation are maintained. The text input from the user is analyzed by a context-based reasoning (CBR) module [66]. There are three main parts to language or dialogue processing: (1) an init/exit module that acts as a bridge between humans and CA; (2) an analysis module that parses conversation text; and (3) an input/output module that takes in new dialogue and returns processed text [67]. The input/output module operates by exploiting the counselling knowledge dictionaries in order to provide replies that are appropriate for the present setting. To enhance the empathic abilities of the robot, the Japanese concept of "unazuki" (nodding), which is considered to be a higher level of empathy representation, is employed on a Raspberry pi based robot. The system is then evaluated on 12 participants, who are given a 15 item Likert scale questionnaire.

In a follow-up study [68], Kurashige et al. performed further experimentation on the comparison of nodding and non-nodding robots and their empathic abilities as per the participants. RNNs were exploited for the text prediction task, whereas rest of the framework was same as [65]. However, the text input from keyboard was replaced by voice input.

## C. State-of-the-art in Visual/Multi-modal Data

*1) Transformers and other DNN based Techniques:*

Xie et al. [69] incorporate the Russel's circumplex model [70] of affect to label the valence (positive or negative emotion) and arousal (intensity of an emotion) values along the sentiment dimensions for the emotion classification module of their proposed empathic robot. Speech recognition was used to detect the emotional states of the user, which is done by exploiting both audio and text data from the speech. GPT is used for pre-training the language model, whereas, for affect analysis, a Support Vector regression based model is used, inspired by [71]. The model is trained on the Empathetic Dialogues dataset. Next-sentence prediction loss, language-modeling loss, emotion classification loss, and valence and arousal regression loss are the four loss functions optimised during fine-tuning. Perplexity (PPL), Emotional Accuracy (EMO ACC), Arousal Mean Squared Error (A. MSE), and Valence Mean Squared Error (V. MSE) were utilised as evaluation metrics (V. MSE).

Tan et al. [26] proposed a multi-modal LSTM with feature-level fusion and local attention that predicts empathic responses from audio, text, and visual features. They use the OMG-empathy dataset from the OMG-Empathy challenge to evaluate their model's ability to recognize empathic responses. For feature extraction from text, YouTube's automated subtitling was utilised to get transcripts and start/end times of each utterances. The GloVe word embeddings [72] were used to extract features per utterance by averaging across the embeddings of the words comprising the speech. For feature extraction from audio data, they extracted 990 low-level acoustic characteristics using openSMILE v2.3.0 [73] and the accompanying emobase configuration file. As for the video features, they retrieved fully-connected feature embeddings from the pre-trained VGG Face CNN models [74]. The features were retrieved from the listener's face for each frame, as they observed that adding the speaker/actor's facial features was not helping the model's recognition performance. They adopted various models and parameters when working with different modalities i.e., text only, text and visual only. For evaluation, they utilise the Concordance Correlation Coefficient (CCC) to compare the model's predictions with participants' own reports of their levels of empathy.

A virtual empathic robot assistant was presented by Fung et al. [25]. They used facial, speech and audio data to perform emotion recognition followed by empathic responses by the proposed virtual agent. They used a CNN for the classification task, where each word was treated as a vector input, while for audio, each frame was used as a vector. A Deep Neural Network (DNN) was suggested for computing the emission probabilities of speech data. At the same time, a Long Short-Term Memory (LSTM) was presented to manage the previous context of each given conversation. They make use of the Kaldi speech recognition tools in order to train acoustic models. A dataset was created from the TED-LIUM corpus release 2 [75] for testing. Six different emotional categories were chosen for the purpose of the experiment. As a baseline for feature extraction, a linear-kernel SVM model from the LibSVM package [76] was utilized in conjunction with the INTERSPEECH 2009 emotion feature set [77] retrieved using openSMILE [78]. For additional experimentation (testing), a new corpus was created using scenes from the popular English sitcoms - Friends and Seinfeld. After this, cross-evaluation was performed using two databases for training and one for testing.

Mathur et al. [79] proposed an automated method for determining whether or not a user has empathized with a robot storyteller, based on the user's eye gaze, facial action units, facial landmarks, head altitude, and point distribution parameters. They also provide a dataset of visual information gleaned from human-robot narrative exchanges in order to promote empathy. Details about the dataset are given in Section 4 (Storyteller robot dataset). Moreover, they experimented with ten different ML models (8 classic ML models and 2 DL models) to detect empathy, where detection of empathy is referred to as predicting the participants' empathic/non-empathic responses. The 8 classical machine learning models were, adaptive boosting

(AdaBoost), bagging, decision trees, linear-kernel support vector machine (Linear SVM), logistic regression, random forest, rbf-kernel support vector machine (RBF SVM), and XGBoost. The two deep learning techniques used were LSTMs and Temporal Convolutional Networks (TCN). Information from eye gaze, face attributes, and head movement at each visual frame was extracted to record participants' visual actions while they listened to the robot storyteller. Eye-gaze directions, the strength and existence of 17 facial action units (FAUs), facial landmarks, head position coordinates, and PDM parameters for face location, scale, rotation, and deformation were all retrieved using the OpenFace 2.2.0 toolkit. To understand and exploit temporal patterns in visual signals that are predictive of empathy, they employed raw sequences of visual data, for deep learning models. 5-fold stratified cross validation was repeated 10 times giving a total of 50 folds. Evaluation was performed using four metrics, namely ACC, AUC, Precision and recall.

Carolis et al. [21] implemented affective reasoning on the NAO robot for simulating empathic behaviors in the context of Ambient Assistive Living (AAL). In their emotion detection architecture, firstly, facial feature extraction is performed using the famous Viola-Jones detector. After this, the Staked Active Shape Model (STASM) method is used to find facial key points. This method uses the Active Shape Model with a simplified version of SIFT descriptors and Multivariate Adaptive Regression Splines (MARS) to match descriptors. For speech recognition, an online service called VOCE (Voice Classifier of Emotions) is employed using a kNN classifier. This tool identifies the valence and arousal values of the given audio. For collection of data, two caregivers, who were looking after two elderly people were asked to record their experience for a period of 9 months. The caregivers were asked to keep a paper journal in which they recorded the day's activities and any noteworthy occurrences, with a focus on how they made them feel i.e., sad, depressed, excited. Empathy was modelled in the robot using Deep Belief Networks (DBN). The robot has a predetermined set of empathic goals, such as comforting the subject to make them feel cared for/loved. Constant monitoring of the audio and facial cues coming from the user are processed through the DBN and a certain empathic goal (e.g. comforting the subject through dialogue) is activated. For evaluation, two methods were adapted: expert-based evalua- tion and user study. No computational/autonomous evaluation techniques were used.

In order to attain more richly-detailed HRI engagements, Bagheri et al. [80] proposed the Automatic Cognitive Empathy Model (ACEM). A stacked autoencoder network, trained and tested on the RAVDESS dataset, is used to identify users' emotional states [81]. The proposed model consists of three distinct modules: an emotion detection module, a perspective taking module, and an empathic behaviour provider module. The implementation of a stacked autoencoder with a softmax activation function is utilised for the purpose of extracting and classifying facial features. While the perspective taking module has been introduced, its implementation has not yet been carried out. The module responsible for providing empathic behaviour encompasses two distinct categories of empathetic responses, precisely parallel and reactive. The parallel empathy approach is a technique employed to mirror the emotional state of the user, thereby fostering a sense of comprehension and validation. In contrast, the reactive technique instructs the robot to respond favourably to the user's emotional condition by making upbeat comments or providing other positive feedback. Classification Rate (CR), False Alarm Rate (FAR), and Confusion Matrix (CM), inspired by [82], are utilised to evaluate the performance of the proposed emotion classification technique. For the evaluation of empathic behavior provider module, 40 participants of mixed gender and personality types were selected. Each participant viewed six videos (representing various emotion classes) while being interacted with by a Pepper robot in order to test the effectiveness of the proposed ACEM. After the experiment was over, each participant filled out three surveys to share their thoughts: the UTAUT questionnaire [83], the friendship questionnaire [9], and the engagement parameter [84]. All of them reflect various facets of the robot's character and are measured on likert scales. For further validation of the results obtained from the evaluation surveys, the Cronbach's Alpha and Wilcoxon Test are applied. A detailed analysis from the results is discussed against various aspects of the robot's empathic and personality characteristics.

*2) Other Techniques:*

Filho et al. [85] proposed RegressionWiSARD and ClusRegressionWiSARD n-tuple regressors and their ensembles in order to predict empathy. They perform experiments on visual and audio data from the OMG-Empathy dataset. For preprocessing of the image data, Adaptive Gaussian filter [86], Sauvola method [87], Canny border detector [88], and Otsu's binarization [89] were adapted. On the other hand, for preprocessing the audio data, mel-frequency cepstral coefficient extraction [90] was implemented, that converts audio to text. A combination of Regression WiSARD (ReW) [91] and ClusWiSARD [92] i.e., ClusRegression WiSARD (CReW) was employed. Following this, ensemble of ReWs and CReWs was formed using three different techniques (Bagging; where each weak learner is trained using a portion of the training data, with replacement. Boost; where weak learners are trained the same as bagging method except for replacement. Naive; where all the weak learners are trained on the whole dataset.) OMG-Empathy Dataset was used to train and test their proposed technique, while CCC was used for validation of each model's performance.

Rasool et al. [10] proposed an HRI based computational emotion model where the internal emotions are defined using psychological studies and generated on 2D (pleasure-arousal) scaling model, whereas, fuzzy logic is used to calculate the intensity of each emotion. The process can be broken down into three primary stages: perception, assessment, and the expression of empathy. Facial expression recognition algorithms [93], based on Constrained Local Model (CLM) with LeaderP clustering algorithms [94] and topological Gaussian

Adaptive Resonance theory algorithm (TGART) are exploited. Face detection is performed using the famous Viola-Jones algorithm. Additionally, Point Distribution Model (PDM) is applied to generate the 2D feature point positions of each patch. While empathy is typically a reflection of two factors i.e., personality and mood, in this work, only the mood is used as the deciding factor while personality remained constant. To express empathy, a virtual facial expression simulator called Grimace is used. Evaluation was performed using facial data to detect the level of empathy as '0' for no-empathy and '1' for empathy.

### D. Reinforcement Learning for AE

As previously mentioned, Reinforcement Learning (RL) is a modern technique that is increasingly used to train artificial agents on unfamiliar data and infuse them with the motivation to explore alternative input and output combinations. Following are some of the works that have incorporated RL for AE.

Leite et al. [95] proposed a multi-modal system for modelling empathy that combines visual and task-related features. A robot i.e., iCat was employed to play a game of chess with children and display empathic responses, such as facial cues and motivating comments during the game. The robot is based on the Open Platform for Personal Robotics (OPPR) software that allows the iCat to be programmed as per the needs of the task. They follow an approach based on RL, so that the robot can learn by reward and penalty the best strategies for a particular user, and adapt to its empathic behaviour accordingly. A questionnaire with three subscales - help, engagement, and self-validation was used to get evaluation from the participants.

In a later work, Bagheri et al. [28] proposed implementation of cognitive aspect of empathy on a robot using interpersonal goals, such as aggression and social behaviour, as well as intra-personal goals, such as parallel emotion and empathic concern. Three modules comprise their proposed framework: Emotion Detection, Reinforcement Learning, and Empathic Behavior Provider. The Emotion Detection module uses a facial emotion detection model to categorise the user's facial expressions into six categories, and the Reinforcement Learning module uses contextual bandit to learn the 'optimal action-selection policy' that enables the robot to select the most appropriate empathic behaviour. After each action, the Q-table is updated and initialized with zeros. The Empathic Behavior Provider module applies the specified behaviours to the robot so that it may respond to the emotions of the user. The proposed framework was implemented on a real-world robot, Pepper. The robot engaged with human participants in a game-playing scenario, and the model was evaluated by asking individuals to report their interactions with the robot on three different scales: friendship, UTAUT, and engagement questionnaires.

Sorrentino et al. [27] used an online platform to train a DRL algorithm - DQN, that used rewards from online participants against several generated facial expressions (emotions) as motivation, in an effort to implement 'affective' empathy. The network used a standard E-greedy Q-learning approach.

Expressions generated by the network were rewarded as 'coherent/incoherent' by the users. The trained network was then implemented on a real robot called CloudIA. A setting of three modes of conversations that featured small conversations, watching a video, and playing a guessing game, was used for the robot to engage with participants. Towards the end of each interactions, Godspeed questionnaire was filled out by each participant to rate the robot's abilities against the 5 sub-scales of the questionnaire. The experiment concluded higher emotion expression capabilities of the robot, but not the 'empathetic behaviour'. Moreover, all the participants involved in the experiment were from the same age group.

Studies that use DRL to implement AE are rather limited in number, at the moment. Since empathy is a complex emotion, it is safe to say that Q-learning is the favourable technique to go with, as it allows exploration of hidden states.

### E. Reflection

One problem with the current research is that most of the experimental settings are monotonous and controlled, making them highly unlikely to be generalized. Also, different types of scales have been used to report user's evaluation making it hard to draw a statistical comparison between these techniques. This is to say that the evaluation should be performed on a general scale that allows for a performance comparison against other works allowing the researchers to improve their future efforts.

Similar is the case with datasets. In addition to scarcity of visual empathy datasets, description of empathy varies in each corpus. It is needless to say that benchmarking requires a consistent modelling of the aspect being evaluated, in order to draw a reliable evaluation. For instance, EPITOME by Sharma et al. uses emotional reactions, interpretations and explorations as three different mechanisms to recognize empathy, whereas, IEMPATHIZE by Hosseini et al. presents two directions of empathy i..e., seeking or providing as the classifying mechanism.

Since Liu et al. [96] proved the inability of BLEU score to validate the performance of dialogue generation tasks, as it does not strongly associate with human judgement, it seems illogical to use this metric for evaluating empathy. This questions evaluation methods of works such as Sharma et al. [33] and Harilal et al. [52]. Table III compares the most widely used autonomous metrics for evaluation of AE.

## III. REVIEW OF EVALUATION METHODS FOR EMPATHY WITH RESPECT TO ARTIFICIAL AGENTS

In this section, we start by reporting our observations on the various empathy assessment scales used in human-robot interactions. We outline the primary features of each scale and their respective application in studies. Additionally, we assess the distinctions between these metrics, particularly in terms of empathy type and other relevant characteristics.

### A. Observations

When it comes to the adaption of a generalised scale for the evaluation of empathy, there are several factors that might lead to ambiguities and variations. One of the reasons is the variations in the interaction levels and application goals of the various artificial agents. For example, a companion robot in an aged care setting may require more compassion, whereas an industrial robot may concentrate on getting the maximum performance. One further thing to be concerned about is how the agents might differ in their features and non-functional aspects including aesthetics, physicality, language skills, and reaction time [97]. The vast majority of the currently available measures are concerned with empathy rather than AE. Because of this, it is more challenging to use them in situations involving HRI-based empathetic exchanges. In addition, the few quantifiers that are applicable to AE need for an assessment based on human subjects. This results in issues, such as the possibility of participants having a bias, the expense in terms of both time and money, and variations in how different people perceive different events.

Several aspects, including user-related factors, context-related factors, and system-related factors, might influence the evaluation of empathy in artificial agents and must be evaluated and addressed. Examples of these factors include, personality, user satisfaction and acceptability of the agent. For an empathy scale to accurately measure the empathetic abilities of an artificial agent, it is of the high significance that it takes into account these aspects [98] [99].

While there are recent works that have tried to adapt human-human interaction (HHI) empathy metrics into the HRI domain, such as RoPE [100] and QMAE [99], validation of these scales is required to ensure the effectiveness and inter-correlation of the items included. This is only achievable if future research includes a validation of the current AE evaluation scales. Another idea is to bring together the characteristics of autonomous and human evaluation metrics together and design a framework that can incorporate the strengths of both. A summary of the scales for empathy evaluation is provided in Table 3.

### B. Godspeed Questionnaire

This popular scale was presented by Bartneck et al. [110] as a series of questionnaires to measure the user's perception of robots. It combines five consistent and validated questionnaires based on 5-point semantic differential scales as a standardized metric for the five key concepts in HRI.

1) Anthropomorphism: rates the user's impression of the robot on five semantic differentials.
2) Animacy: rates the user's impression of the robot on six semantic differentials.
3) Likeability: rates the user's impression of the robot on five semantic differentials.
4) Perceived Intelligence: rates the user's impression of the robot on five semantic differentials
5) Perceived Safety: rates the user's emotional state on three semantic differentials.

Implementation of the Godspeed questionnaire can be seen by Johanson et al. [115], who implement expressions of humor on a robot in healthcare setting to gauge its impact on users.

### C. Barrett-Lennard Relationship Inventory (BLRI)

This scale was proposed by Barrett-Lennard et al. [101], initially as an instrument to measure the emotional relation between humans. However, it can be used to indicate human-robot relations as well. The BLRI consists of 64 items (16 to measure each of the four dimensions). The dimensions/sub-scales were as follows: empathic understanding, level of regard, un-conditionality of regard, and congruence. Item wording reflects either positive (e.g., "She or he understands me.") or negative ways of responding to a person. Subjects used 6-point scales (3 = strongly feel that it is true to -3 = strongly feel that it is not true) to respond to items.

### D. Davis' IRI

Davis' IRI [4] for evaluating empathic relations is one of the most popular empathy scales of all time. It is a 28-item self-report questionnaire with four 7-item subscales, each assessing a different component of empathy. These subscales represent characteristics that are important in interpreting the aspects of empathy in any individual. Following are the subscales of the Davis' scale with a brief overview.

1) Perspective Taking (PT) scale - It evaluates the capacity to see ordinary events through the perspective of the attitudes held by other individuals.
2) Fantasy (FS) scale - It assesses a person's propensity to project their own thoughts and feelings onto the emotions and actions of fictional characters that they encounter in works of fiction such as novels, films, and plays.
3) Empathic Concern (EC) scale - It assesses the likelihood of having sentiments of warmth, compassion, and care for other people in everyday life.
4) Personal Distress (PD) scale - It examines usual emotional responses, but instead of worry for others, it delves into one's own sentiments of personal disquiet and discomfort in response to the feelings of somebody else.

### E. Robot's Perceived Empathy (RoPE) scale

There are a total of 18 questions that make up the Robot's Perceived Empathy (RoPE) scale [116], with two sub-categories, measuring either "empathetic understanding or empathic response". Each sub-scale's items were chosen with consideration for research on the efficacy of empathy in HRI settings [117]. Not all of the questions on the original empathy survey applied to robots, hence some were removed. Implementation of the RoPe scale can be seen in the work by Daher et al. [118].

### F. State Empathy Questionnaire

The self-report Emotion Awareness Questionnaire (EAQ) [109] was made to reflect the key ideas of emotional awareness. It has six scales: the ability to tell the difference between emotions and find out where they came from (Differentiating Emotions); paying attention to the physical aspects of the emotion experience (Bodily Awareness, i.e. being aware that emotions are accompanied by physical symptoms; the ability to talk about emotions (Verbal Sharing); the blunt expression of emotions (Acting Out); and others' feelings (Attend to Others' Emotions and Emotional Analyses, respectively). The initial scale of 40 points has been decreased to 30, and irrelevant items have been excluded from the examination of cognitive and emotional empathy.

TABLE IV
SUMMARY OF EVALUATION SCALES FOR EMPATHY.

| Study | Method | Components/Subscales | Associated (originally) with HRI? | Addresses Empathy Directly? | Year |
|---|---|---|---|---|---|
| Barrett-Lennard et al. [101] | Barrett-Lennard Relationship Inventory (BLRI) | Empathic understanding, Level of regard, Un-conditionality of regard, Congruence. | No | Partially (cognitive aspects) | 1962 |
| R Hogan [102] | Hogan's Empathy Scale | California Psychological Inventory Minnesota Multiphasic Personality Inventory, Chapin Social Insight test | No | Yes Cognitive (role-taking) | 1969 |
| Mehrabian & Epstein [103] | QMEE/EETS | Susceptibility to Emotional Contagion, Appreciation of the Feelings of Unfamiliar and Distant Others, Extreme Emotional Responsiveness, Tendency to Be Moved by Others' Positive Emotional Experiences, Tendency To Be Moved by Others' Negative Emotional Experiences, Sympathetic Tendency, and Willingness to Be in Contact with Others Who Have Problems. | No | Yes (Emotional) | 1972 |
| Batson et al. [104] | Emotional Response Questionnaire (ERQ) | Personal distress, Empathy/Sympathy | No | Yes (Emotional) | 1982 |
| M H Davis [4] | Interpersonal Reactivity Index (IRI) | Perspective taking, Fantasy scale, Empathic concern, Personal distress | No | Yes Emotional, Cognitive, (compassion) | 1983 |
| Biggam et al. [105] | Positive and Negative Affect Schedule (PANAS) | Positvie affect, Negative affect | No | Yes (Emotional) | 1996 |
| Caruso et al. [106] | Multidimensional Emotional Empathy Scale (MDEES) | Empathic suffering, Positive sharing, Responsive Crying, Emotional attention, Feeling for others, Emotional contagion | No | Yes (Emotional) | 1998 |
| Baron-Cohen et al. [107] | Empathy Quotient (EQ) | Clinical Empathy, Social distress | No | Yes (clinical) | 2004 |
| Jolliffe et al. [108] | Basic Empathy Scale (BES) | Emotional congruence, Cognitive aspects | No | Yes (Emotional, Cognitive) | 2006 |
| Rieffe et al. [109] | Emotion Awareness Questionnaire | Differentiating emotions, Bodily awareness, Verbal sharing, Acting out emotions, Attending to other's emotions and analysis of own emotions | No | Yes (Emotional, Cognitive) | 2007 |
| Bartneck et al. [110] | Godspeed Questionnaire | Anthropomorphism, Animacy, Likeability, Perceived Intelligence, Perceived Safety | Yes | No | 2009 |
| Spreng et al. [111] | Toronto Empathy Questionnaire (TEQ) | Emotional contagion, Emotional comprehension, Sympathetic physiological arousal, Con-specific altruism | No | Yes (Emotional, Cognitive) | 2009 |
| Rieffe et al. [112] | Empathy Questionnaire for Children and Adolescents (EmQue-CA) | Affective empathy, Cognitive empathy, Intention to comfort | No | Yes (Emotional, Cognitive) | 2010 |
| Shen et al. [113] | State Empathy Questionnaire | Emotive, Cognitive, Associative (ability to relate) | No | Yes (Emotional, Cognitive) | 2010 |
| Reniers et al. [114] | Questionnaire of Cognitive and Affective Empathy (QCAE) | Cognitive empthy, Emotional empathy | No | Yes (Emotional, Cognitive) | 2011 |
| Charrier et al. [100] | Robot's Perceived Empathy (RoPE) | Empathic understanding, Empathic response, Filler items (for HRI) | Yes | Yes (perceived) | 2019 |
| Putta et al. [99] | QMAE | Empathic Understanding, Empathic response, Empathic relationship | Yes | Yes (perceived) | 2022 |
| Leite et al. [95] | Friendship Questionnaire | Intimacy, Emotional security, Social presence | Yes | No | 2006 |

### G. Emotion Awareness Questionnaire

The self-report Emotion Awareness Questionnaire (EAQ) [109] was made to reflect the key ideas of emotional awareness. It has six scales: the ability to tell the difference between emotions and find out where they came from (Differentiating Emotions); paying attention to the physical aspects of the emotion experience (Bodily Awareness, i.e. being aware that emotions are accompanied by physical symptoms; the ability to talk about emotions (Verbal Sharing); the blunt expression of emotions (Acting Out); and others' feelings (Attend to Others' Emotions and Emotional Analyses, respectively). The initial scale of 40 points has been decreased to 30, and irrelevant items have been excluded from the examination of cognitive and emotional empathy.

### H. Emotional Response Questionnaire (ERQ)

There are various versions and types of Emotional Response Questionnaires used by different researchers from time-to-time. It has been updated and changed as per the requirement of each study where it was deployed to, which is why there is no single description for this certain empathy scale, however, the one most frequently used structure of ERQ was proposed by Batson et al. [104]. Following are some of its features. The ERQ was a list of 28 adjectives that described how people felt. Eight of these adjectives (alarmed, grieved, upset, worried, disturbed, distressed, troubled, and perturbed) were found in previous research to describe feelings of personal distress, and six others were found to describe feelings of empathy (sympathetic, moved, compassionate, warm, softhearted, tender). Respondents were asked to rate how much of each emotion they were feeling while watching the worker, on a 7-point scale (1 = not at all, 7 = extremely).

### I. Questionnaire of Cognitive and Affective Empathy (QCAE)

QCAE was developed and validated by Reniers et al. [114]. The research compiled a total of 65 items for the purpose of gauging cognitive (29 items) and affective (36 items) empathy. These items were derived from the Empathy Quotient, the Hogan Empathy Scale, the Empathy subscale of the Impulsiveness-Venturesomeness-Empathy Inventory, and the IRI. The Hogan Empathy Scale was used to measure cognitive empathy. The QCAE is a reliable instrument for evaluating both cognitive and emotional aspects of empathy.

### J. Positive and Negative Affect Schedule (PANAS)

PANAS [105] is a likert scale of five points that has been devised to capture the respondents' sentiments as they reported their experiences. The scale ranges from 1 (very little or not at all) to 5 (extremely). The level of participants' ability to empathise with others was evaluated by calculating the absolute difference in PANAS emotion scores between those reported by the targets in the clips and those reported by the participants themselves. The PANAS scales are valid, reliable, and independent measures of both positive and negative affect, regardless of the population investigated, the time period examined, or the answer format used.

### K. Multidimensional Emotional Empathy Scale (MDEES)

Proposed by Caruso et al. [106], the MDEE scale has a total of 30 items (questions), and it was originally administered to a sample size of 793 adults and children. Principal Component Analysis (PCA) was used to get a total of six significant variables out of the data. On the basis of these variables, sub-scales are formed. This scale assesses the emotional components of empathy and can be used to evaluate emotional empathy. Additionally, it provides specific sub-scales for further analysing the results.

### L. Hogan's Empathy Scale (HES)

R Hogan [102] devised the HES in 1969. After assigning a criteria for rating empathy, the first step in developing Hogan's empathy scale was to compare the answers of 57 men who had high ratings for empathy against 57 men who had low ratings for empathy across the combined-item pools of the California Psychological Inventory (CPI). Each rating scale has four questions that follow the Likert scale and have seven levels, with opposing adjectives serving as anchors at each end. One of the items that makes up the Adjustment scale, for example, is anxious-calm.

### M. Questionnaire Measure of Emotional Empathy (QMEE) & Emotional Empathic Tendency Scale (EETS)

QMEE and Emotional Empathic Tendency Scale are more or less the same, and are both presented in the same study by Mehrabian and Epstein [103]. The QMEE is a questionnaire that consists of 33 items, each of which the candidate rates on a scale ranging from very strong disagreement (-4) to very strong agreement (+4). The signs that came before negative items are changed, and the sum of their scores on all 33 questions is used to get the respondent's overall score. Therefore, a high score indicates a high level of empathy. The EETS, same as the QMEE, has 33 questions on your propensity to feel other people's emotions. For instance, "it's hard for me to understand why certain things bother people so much, and it saddens me when I see a lone stranger within a bunch". There is a 9-point scale from -4 (very strong disagreement) to +4 (strong agreement) that respondents use to score each statement (very strong agreement).

### N. Empathy Quotient (EQ)

Empathy Quotient by Cohen et al. [107] measures empathy on a 40-item Likert scale for empathy and a 20-item scale for controls. The greatest possible score on the (EQ) is 80, with a minimum score of 0. Scores on the individual empathy items may be 2, 1, or 0, giving the scale a total range of 20 to 80. The filler questions are not related to empathy and are there just to make sure that a participant does not get overwhelmed by intense focus on empathy. The EQ was designed to be short, easy to use, and easy to score. An example item from the scale is "It is hard for me to see why some things upset people so much." Participants have the option to slightly/strongly agree/disagree, scoring 1/2 points, respectively.

O. Empathy Questionnaire for Children and Adolescents (EmQue-CA)

The EmQue-CA consists of 21 items generated by the Rieffe et al. [112]. There were three scales: (1) Affective empathy (nine items, like "When a friend is upset, I feel upset too") measures how much one shares another person's feelings. (2) Cognitive empathy (six items, like "If a friend cries, I often understand what has happened") measures how much one understands why another person is upset. (3) Intention to comfort (six items, like "If a friend is sad, I want to do something to make it better") measures how much one wants to help someone who is upset On a scale from one to three, participants were asked to indicate whether or not they found the description to be true for them: (1) not true, (2) somewhat true, and (3) true. Re-scoring was performed on all of the questions such that higher scores would imply a stronger level of empathy.

P. The Toronto Empathy Questionnaire (TEQ)

Proposed by Spreng et al. [111], the TEQ contains 16 questions that encompass a wide range of attributes associated with the theoretical facets of empathy. The affective aspect of empathic responding is thought to be related to such phenomena as emotional contagion, emotion comprehension, sympathetic physiological arousal and con-specific altruism, all of which are represented in TEQ items. The TEQ correlates highly with Davis' IRI, however, it is an amalgamation of several empathy scales, i.e, Hogan's Empathy Scale, QMEE, Balanced Emotional Empathy Scale , Jefferson Scale of Physician Empathy and a few more.

Q. Basic Empathy Scale (BES)

Presenting the BES, Jolliffe et al. [108] also mentioned that the definitions of empathy used for the development of the QMEE and IRI, and the items on these scales, may be failing to measure empathy adequately. To clarify the definition of empathy and segregate its types on a better level, BES was presented, with an updated concept of empathy, especially cognitive empathy, since the previous questionnaires were not able to measure cognitive empathy. The identification and development of affective empathy questions, which measured emotional congruence, is a primary focus of this scale.

R. Questionnaire to Evaluate Empathy in Artificial Agents (QMAE)

The QMAE [99] is one of the few empathy evaluation methods specifically designed for use in human resources information systems. It's a tweak to the RoPE scale that takes into account some of the components of AE that were left out of the RoPE scale. It incorporates a few aspects from some of the popular questionnaires, such as the BLRI scale [101], Godspeed questionnaire [110], Companionship Scale for Artificial Pets [119], and AttrakDiff questionnaire [120]. It is a 6 point likert scale with values ranging from -3 to 3. The degree to which a user feels connected to a digital assistant is quantified here. The relationship depends on how the user and artificial agent interact, how the user responds to the artificial agent, how accurate the agent's predictions are, and how the user interprets the results.

S. Empathy Assessment Index (EAI)

Proposed by Lietz et al. [121], the EAI is a 20-item likert scale-based self report empathy questionnaire. It has four sub-scales that are namely, affective response, self-other awareness, perspective taking and emotion regulation. The scale was developed to measure human empathy, but it may be modified for use with artificial agents if required; it already has components like self-other awareness and perspective taking that are important for medical support robots.

T. EMOTE Questionnaire

Emote questionnaire is a part of the EMOTE project [122]. It is a 14 item self-report scale, whose items are mostly inspired by the IRI [4]. At the same time, it is also one of the few questionnaires that have been used to evaluate behaviour in robots i.e., social assistive robots (SARs). However, their are studies that have used the EMOTE questionnaire with a few modifications (removing irrelevant items) for evaluation of emotion in robots [123].

U. Friendship Questionnaire

Inspired by Leite et al. [95], the friendship questionnaire is a five-point likert scale used to evaluate the friendly characteristics of an artificial agent. It is important to include this in the empathy related works because investigating friendship functions including closeness, emotional security, and social presence can provide some indications of improvement in human-robot relationship [80].

V. Reflection

Human-based evaluation methods are widely regarded as the most reliable and accurate. These methods enable evaluators to make nuanced and context-dependent judgements of empathy, which is essential for understanding the complex nature of human social interactions. As such, human-based evaluation methods are often considered the gold standard for empathy evaluation [114] [100].

On the other hand, the current research indicates an increasing inclination towards the creation of autonomous techniques for assessing empathy, specifically in the domain of natural language processing (NLP) and affective computing [33]. These techniques exhibit the capability to automate the evalu- ation of empathy on a large scale and to furnish awareness into the affective aspects of large datasets. Moreover, utilization of traditional scales such as the EAI [121] can prove useful in formation of an automated system for empathy evaluation in empathy in artificial agents since they already include elements such as self-other awareness.

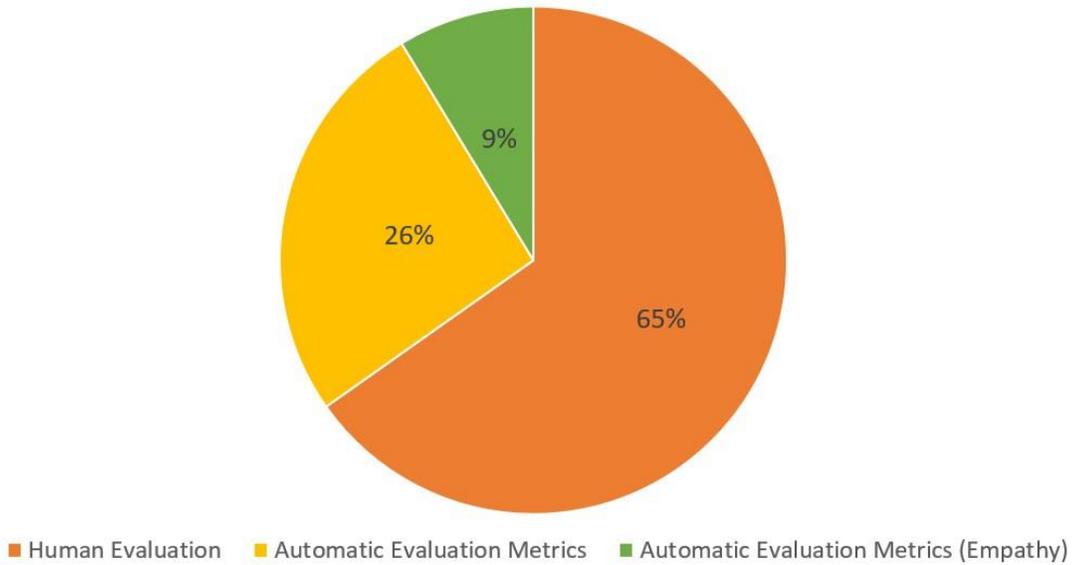

Fig. 5. Comparison of frequency of different types of evaluation metrics used for AE [13]

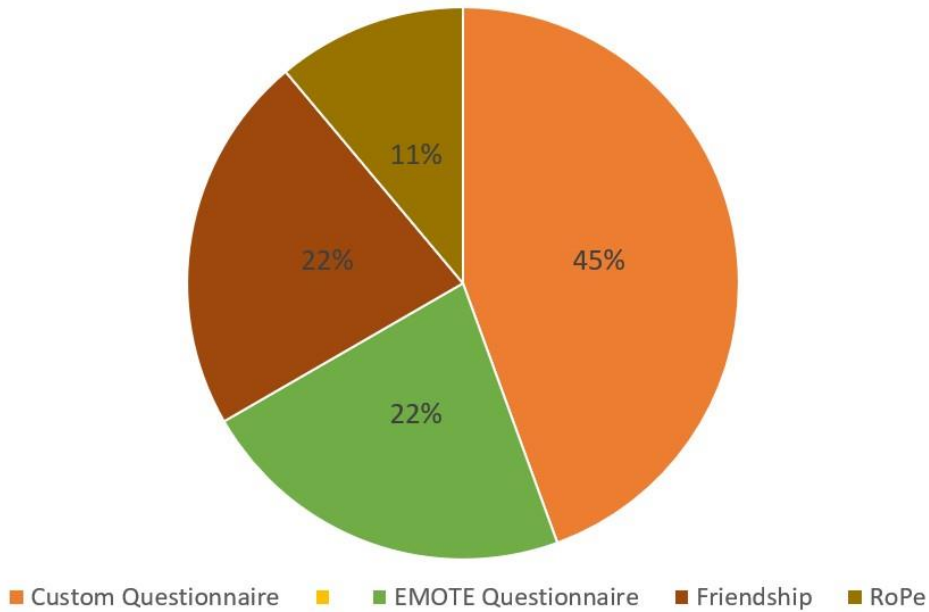

Fig. 6. Comparison of frequency of various human evaluation metrics used for AE [13]

While there are fewer studies that have shown interest in development of autonomous empathy evaluation metrics, there has not been any works conducted to authenticate these metrics. Further research is required to establish the efficacy of these metrics and to ascertain their constraints for AE evaluation. A visual depiction of the ratio of usage among different types of evaluation metrics can be seen in Figures 5, 6 and 7. The objective behind this is to emphasise the prevalent scales utilised in contemporary research on AE.

## IV. REVIEW OF DATASETS FOR ARTIFICIAL EMPATHY

### A. Observations

Most of the visual datasets, such as OMG-Empathy, use a controlled environment, where listeners and speakers are given a set of dialogues. There are drawbacks to maintaining a controlled setting, such as a disconnect between the speakers' scripted narratives and the way they would communicate genuine narratives. Furthermore, these datasets do not include information about the effect of the annotated behavior as to how the listeners perceived it. On the other hand, due to the fact that the text-based datasets have often been taken from random talks on various online platforms such as Facebook and other online blogs, their raw nature is superior to that of the visual ones in this context. It will be useful to have visual datasets collected in their natural settings since this will make it possible to train artificial agents, particularly robots, how to interact and interpret dialogue in a manner that is more natural.

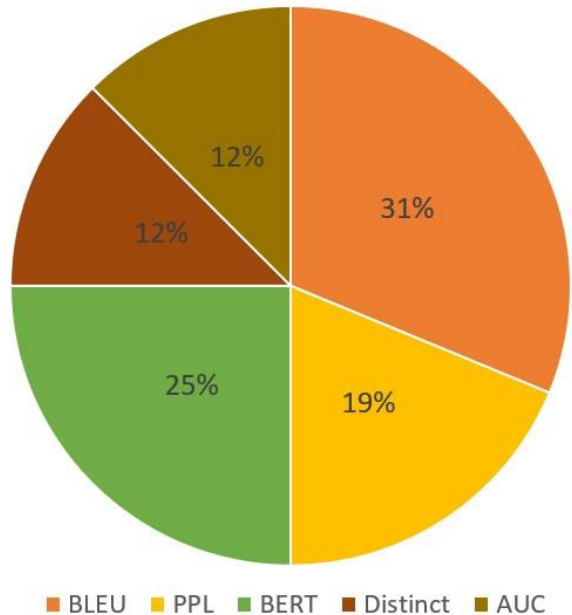

Fig. 7. Comparison of frequency of various autonomous evaluation metrics used for AE [13]

Furthermore, classification of empathy is binary in the existing visual datasets, which hinders developing an understanding of extent/degree of empathy in the artificial agents. Empathy is a complex emotion, and it should not be limited to a set of defined emotions. For instance, a human analyses the situation by verbal and other cues, takes into account the overall context and valence of the dialogue, connects it with previous experiences, and then responds.

Another significant challenge with some of the existing empathy datasets is their underutilization, primarily due to their relatively small data size. Researchers often prefer to employ customized datasets, which may introduce uncertainties, as empathy is multifaceted and diverse in nature. Moreover, several of these datasets exclusively capture an individual's emotional state, without considering the accompanying reactions of the other person and appropriate responses. In the following we discuss empathy evaluation datasets.

### B. Facebook Empathic Dialogues (FED)

FED [34] is a public dataset consisting of 25,000 discussions that are based on different emotional scenarios. Each conversation is rooted in a particular circumstance where a speaker was experiencing a certain emotion and a listener was reacting to their experiences. The dataset is comprised of crowdsourced one-on-one chats, and it addresses a diverse range of emotions while maintaining a sense of equilibrium. The talks are collected from 810 different individuals and are made accessible to the public under the framework of ParlAI3. Every interaction has been partitioned into around 80% train, 10% validation, and 10% test. The data is divided in such a manner that all sets of conversation in which the same speaker delivers the initial scenario description will be included inside the same partition. This is done to avoid the discussion of the same situation being repeated in multiple partitions. The total number of chats for the final training, validation, and testing splits are 19533, 2770, and 2547, respectively. Adaptation of the dataset can be seen in [55].

### C. OMG-Empathy

The OMG-empathy [124] dataset was created by capturing the audio and visual data from a live conversational encounter between two speakers and a listener, in which the speakers and the listener were sitting in front of each other. The speakers and the listener were facing each other. In each of the scenarios, there are two different speakers, and each of them tells two different tales. After each recording, the participants were given the opportunity to rewatch the interactions on a computer screen and were prompted to make notes regarding the manner in which the interaction influenced their affective state in terms of valence using a continuous scale with values ranging from positive one (1) to negative one (-1). There are two distinct protocol settings: personalised and generic. The dataset includes three pre-defined types of separation sets:

TABLE V
SUMMARY OF DATASETS FOR AE.

| Study | Dataset | Modality | Source | Year |
|---|---|---|---|---|
| Rashkin et al. | Empathetic Dialogues | Textual | Facebook conversations | 2018 |
| Sharma et al. | EPITOME | Textual | Reddit (mental health) | 2020 |
| Amanova et al. | Daily Dialogue | Textual | AMI, MapTask, SWBD | 2016 |
| Welivita et al. | EmotionalDialogues in OpenSubtitles (EDOS) | Textual | Open Subtitles | 2021 |
| Mathur et al. | Storyteller Robot Dataset | Textual | Stories | 2021 |
| Hosseini et al. | IEMPATHIZE | Textual | Online (cancer discussion) forums | 2021 |
| Harilal et al. | Medical Question Answering (MQA) | Textual | eHealth forums | 2020 |
| Barros et al. | OMG-Empathy | Video | Participant recordings | 2019 |
| Chen et al. | Neural Image Commenting with Empathy (NICE) | Image | Images (websites), Comments user generated | 2021 |

training, validation, and testing. These sets are applicable to both methods. The self-assessment annotations are used to divide the samples into training and testing sets, and these sets are then balanced against one another. Out of all the tales, four of them are used for training, one for validation, and the other three for testing. There are ten videos connected to each story, and each listener watched one of them. Adaptation of the dataset can be seen in [125].

### D. Neural Image Commenting with Empathy (NICE)

The Neural Image Commenting Evaluation (NICE) dataset [126] includes approximately two million pictures, 7 million human-generated comments related to those images, and over 28,000 human annotated examples. Following the application of the filters, the dataset now contains a total of 2,150,528 photos as well as 6,720,542 comment dialogue threads. According to the study by Chen et al., the NICE dataset utilises a substantially less number of abstract words than the other datasets, while having the biggest vocabulary size. This indicates that the dataset is capable of producing words and remarks that are easier to comprehend and more cohesive than those produced by any previous empathy datasets.

### E. EPITOME

Epitome [33] is a publicly available dataset that contains 1.6 million posts and 8 million interactions. These are derived from discussions posted on 55 sub-reddits that are focused on mental health (by reddit.com). A subset of 10 thousand exchanges within these threads have been annotated with regards to empathy. Crowd-workers were provided a pair of posts (seeker post and response post) and asked to identify the existence of the three communication channels in EPITOME (emotional reactions, interpretations, and explorations), one at a time. This was done as part of the annotation process. Adaptation of the dataset can be seen in [127].

### F. Medical Question Answering (MQA)

The Medical Question Answering [52] dataset is an accumulation of question-answer pairs that have been scraped from the internet. These discussion pairs consist of medical advice taken from a variety of online medical consulting forums, including eHealth Forum, HealthTap, and WebMD, amongst others. There are 35,294 questions and answers included inside the dataset.

### G. IEMPATHIZE

IEMPATHIZE [37] is a publicly accessible dataset consisting of 5,007 phrases that were taken from an online cancer network and categorised as either seeking empathy, delivering empathy, or having none. The breast and lung cancer discussion forums provide the source material for the chosen sentences. The sentences employed in the dataset all have a maximum length of five words to ensure that the annotation process is as accurate and efficient as possible. The annotation of the sentences into the aforementioned three categories was the responsibility of two graduate students. This dataset is one of a kind because, in addition to just identifying empathy, it also indicates whether a person is seeking or offering empathy to another.

### H. Daily Dialogue

DailyDialogue [128] is a vast dataset consisting of everyday interactions that have been labelled as belonging to one of four distinct categories: inform, questions, directives, or comply. It has 13,118 conversations that are broken up into three sets: a training set with 11,118 dialogues, a validation set with 1000 dialogues, and a test set with 1000 dialogues. There are around 8 speaker turns involved in each debate, and each turn is of 15 tokens. The dataset is annotated based on a classification system that consists of seven basic emotion categories. Three professionals from the domain of dialogue and communication theory performed the annotation of the dataset.

### I. Emotional Dialogues in OpenSubtitles (EDOS)

EmotionalDialogues in OpenSubtitles (EDOS) [48], is a large-scale public dataset that contains 1 million emotional dialogues that have been extracted from the subtitles of movies. It contains dialogues that have been extracted from the website

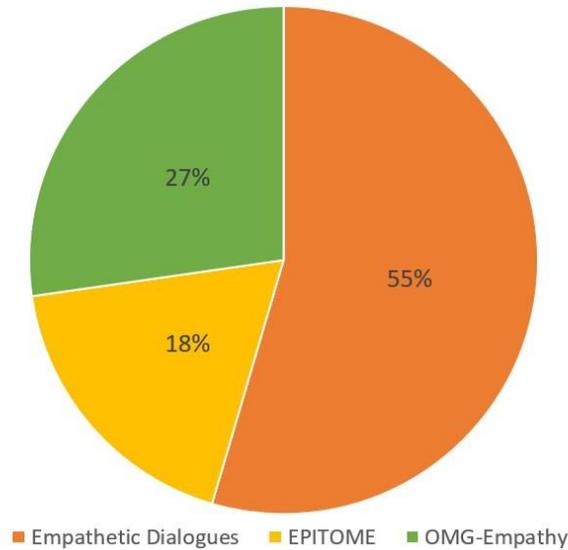

Fig. 8. Comparison of frequently used AE datasets.

Open Subtitles (OS). In this dataset, each dialogue turn is automatically annotated with 32 fine-grained emotions, eight empathic response categories, and a Neutral category. They annotate a portion of the dataset (9k) using a semi-automated manual annotation in conjunction with a low-quality classifier i.e., BERT. After that, this subset is then utilised train an emotion classifier that will be used to automatically label the remaining dataset. It is one of the only two datasets that have all 32 emotion labels, the other being the NICE dataset.

### J. Storyteller Robot Dataset

Mathur et al [79] introduced storyteller robot dataset. To generate this dataset, they have a desktop robot named QT read three distinct stories to a group of people, who then fill out a likert-scale questionnaire to describe their degree of empathy at the conclusion of each story. Subsequently, depending on the participant's answers to the questionnaire, an empathy score (ES) is assigned to each story.

### K. Reflection

Despite the variety of datasets available in the domain, research shows utilization of only a few (including both text and visual). There are several reasons as to why these datasets are favoured over the others. For instance, the Empathetic Dialogues dataset comprises a substantial quantity of conversational interactions that have been annotated with empathy-related labels denoting the presence or absence of empathy. The utilisation of such data enables researchers to effectively train machine learning models in the identification of patterns and characteristics that are linked to empathetic behaviour. Figure 8 shows the ratio between the popularly used AE datasets.

## V. CONCLUSION AND FUTURE RESEARCH PROPOSITIONS

As more and more artificial agents are being used to assist healthcare workers in settings like aged care homes and online mental health platforms, the ability to empathise with their users is more crucial than ever. Many academics have suggested state-of-the-art DL approaches, like Transformers for multi-modal data, to implement artificial empathy, with the hope of evoking empathy in humans by means of artificial agents or vice versa. However, approaches such as reinforcement learning, which have proven quite beneficial with unsupervised and unseen data, have not yet been investigated in this field.

Unlike the seven fundamental emotions such as anger or sadness, empathy is a complex human emotion; thus, it cannot be recognised simply from image or audio data without additional evaluation. In this regard, recent research has shown the use of questionnaires designed specifically to assess the empathic capacities of an artificial agent. It is important to note that, to yet, only two assessment instruments have been established exclusively to evaluate AE. Moreover, none of these two scales have been used by other studies in the field to prove their effectiveness.

Last but not least, in contrast to other tasks using computer vision, artificial empathy is negatively impacted by having a smaller number of visual datasets. Because the majority of contemporary methods depend on a higher quantity of data, this further complicates the process of developing state-of-the-art classification and assessment approaches. Currently, there is a single video-based data set for AE, that too with a restricted degree of empathy. Conversely, text-based systems have made significant progress in recent years because of the availability of more publicly accessible data and methods such as Transformers.

Based on our findings, we propose the following future research directions, which are discussed in detail:

- Is my empathy, your empathy? - Generalized idea of AE
- The more the merrier. - Call for datasets

- Unseen, unpredictable data. - Reinforcement Learning
- Beyond the overused data types. - Non-verbal cues
- Why does my robot empathize like this? - Towards explainability
- Autonomous, but at what cost? - Evaluation metrics

### A. Generalized idea of AE

When designing "empathetic" agents, we are frequently faced with the challenge of deciding how to assess their performance (particularly involving user studies). In the available literature, it can be seen that the idea of artificial empathy differs from one study to another depending on the experimental circumstances. This creates misunderstanding when analyzing the performance of various strategies. Therefore, there ought to be a uniform concept of AE that the scientific community should adhere to for future work.

### B. Call for large-scale datasets

Empathy is the ability to comprehend and react to the emotional states of others, which is difficult for artificial agents to mimic. Large and diverse datasets are required to train DL models in order to increase their capacity to recognise and react to emotions. These datasets should include a diverse variety of samples of various emotional states, as well as information about the context in which they are manifested. Furthermore, additional data in several modalities i.e., video, text and images, is required to increase agents' capacity to react to emotions in an appropriate and human-like way.

### C. Reinforcement Learning

RL methods have been used in a number of experiments examining artificial empathy, however the findings are not conclusive. This is because they have not been evaluated with existing datasets. Therefore, further study is required to comprehend how RL may be utilised successfully to train models for artificial empathy and how to enhance the capacity of RL agents to comprehend and react to emotions in a human-like way. Different motivation/goal settings, as to how the actions/responses of an artificial agent impact the empathy evaluation can help further improve the performance.

Notably, RL is a complex area, and merging it with natural language generation and empathy elicitation in robots, which is necessary for AE, is much more difficult and resource-intensive.

### D. Non-verbal cues

The majority of the current works that deal with visual data concentrate on face expressions. As the emotion recognition task helps analyze, other forms of cues, such as tone of voice and body gestures, may prove to be particularly effective in providing more information about a person's mood and, therefore, the degree of empathy required.

### E. Towards Explainability

Explainable AI (XAI) has lately gained popularity due to its capacity to analyse and explain the reasoning behind the behaviour of various ML techniques. Creating explainable AI models that can offer a clear explanation for their empathic judgements, to assist create trust with users and guarantee that they are making decisions that line with human values, might thus be one of the promising topics to pursue.

### F. Evaluation Metrics

To assess a system's empathic capabilities, many characteristics of empathy must be analysed. However, most studies that employ autonomous evaluation metrics such as F1 or BLEU scores miss out on most elements of empathy and are instead focused on the system's language generation ability. Human evaluation, on the other hand, has a better grasp on empathic evaluation, but at the expense of complications associated with human valuation, such as time constraints. As a result, evolving metrics are required to address additional features of artificial empathy while being cost and time efficient.